\documentclass{article}
\usepackage{spconf,amsmath,graphicx}
\usepackage{amssymb}
\usepackage{tikz}
\usepackage{adjustbox}
\usepackage{multirow}
\usepackage[algo2e,ruled]{algorithm2e}
\usepackage{booktabs}

\DeclareMathOperator*{\argmax}{arg\,max}

\title{Deep Clustering with Concrete $k$-means}

\name{Boyan Gao$^1$, Yongxin Yang$^2$, Henry Gouk$^1$, Timothy M. Hospedales$^{1,3}$}
\address{$^1$School of Informatics, University of Edinburgh, United Kingdom \\
$^2$Centre for Vision, Speech, and Signal Processing, University of Surrey, United Kingdom \\
$^3$Samsung AI Centre, Cambridge, United Kingdom}

\begin{document}
%
\maketitle
\begin{abstract}
We address the problem of simultaneously learning a $k$-means clustering and deep feature representation from unlabelled data, which is of interest due to the potential of deep $k$-means  to outperform traditional two-step feature extraction and shallow-clustering strategies. We achieve this by developing a gradient-estimator for the non-differentiable $k$-means objective via the Gumbel-Softmax reparameterisation trick. In contrast to previous attempts at deep clustering, our concrete $k$-means model can be optimised with respect to the canonical $k$-means objective and is easily trained end-to-end without resorting to alternating optimisation. We demonstrate the efficacy of our method on standard clustering benchmarks.
\end{abstract}
\section{Introduction}
Clustering is a fundamental task in unsupervised machine learning, and one with numerous applications. A key challenge for clustering in practice is the inter-dependence between the chosen representation of the data and the measured distances that drive clustering. For example, the classic and ubiquitous $k$-means algorithm assumes a fixed feature representation, distance metric, and underlying spherical cluster distribution. This set of assumptions leads to poor performance if $k$-means is applied directly to complex high dimensional data such as raw image pixels, where right representation for clustering is a highly non-linear transformation of the input pixels. This observation motivates the vision of end-to-end deep clustering. Joint learning of  data representation and $k$-means clustering has the potential to learn a ``$k$-means friendly'' space in which high-dimensional data can be well clustered, without problem-specific hand-engineering of feature representations. More generally, unifying unsupervised clustering and representation learning has the potential to help alleviate the data annotation bottleneck in the standard supervised deep learning paradigm.

The key challenge with realising this deep clustering vision is the non-differentiability of the discrete cluster assignment in the $k$-means objective. Two recent methods---DEC \cite{xie2016unsupervised} and DCN \cite{yang2017towards}---attempt to address this issue by proposing surrogate losses and alternating optimization heuristics, respectively. However, the surrogate loss used by DEC may not lead to the optimal solution of the $k$-means objective. Furthermore it makes use of soft instance-cluster assignment, which is known to favour overlapping clusters compared to hard assignment methods~\cite{kearns1997information}, and more importantly does not provide the discrete assignments necessary for interpretibility in some applications of $k$-means~\cite{kearns1997information}. In contrast, DCN resorts to alternating optimisation rather than end-to-end gradient-based learning. This is sub-optimal and more importantly restricts the ability to integrate clustering as a module in a larger backprop-driven deep network.  In this paper we propose concrete $k$-means (CKM), the first end-to-end solution to solving the true $k$-means objective jointly with representation learning. We achieve this by adapting the Gumbel-Softmax reparametisation trick \cite{jang2016categorical} to allow differentiation and backpropagation through the discrete cluster assignment. This CKM algorithm enables joint training of cluster centroids and feature representation, and is easy and fast to optimise using standard deep learning optimsation methods. Furthermore, we show that as a byproduct, our CKM also provides a solver for shallow $k$-means with comparable performance to the standard $k$-means++ \cite{arthur2017kmeansplus}.

To summarise, our main contribution is the concrete $k$-means algorithm, the first joint end-to-end solution to the learning of clusters and representations in discrete-assignment deep $k$-means.

\section{Related Work}
Clustering methods aim to find subgroups of data that are related according to some distance metric or notion of density. The performance of distance-based clustering algorithms is highly dependent on the data representation, and the goal of \emph{deep} clustering is to learn a representation of the data that best facilitates clustering. $k$-means is perhaps the most ubiquitous clustering method \cite{lloyd1982least,arthur2017kmeansplus,xu2019powerKmeans}, and it is widely used due to its simplicity, efficacy, and interpretability of its hard cluster assignment. For this reason, several attempts have been made to develop deep $k$-means generalisations. However, this is challenging due to the required hard assignment between data points and cluster centres in the $k$-means objective being hard to reconcile with gradient-based end-to-end learning. Xie et al.~\cite{xie2016unsupervised} show how to jointly optimise an autoencoder and a $k$-means model to get a ``$k$-means friendly'' latent space. The hard assignment in the $k$-means objective prevent them from optimising the true loss function, so their DEC method makes use of an approximation based on soft assignment of instances to clusters. However, this surrogate objective means that the solution to their model is not necessarily a minimum of the $k$-means objective. In contrast, DCN \cite{yang2017towards} resolves the issue by alternating optimisation. Each minibatch of training data is first used to update the deep representation while keeping the centroids held constant, and then used to update the centroids while holding the representation constant. However, alternating optimisation may be slow and ineffective compared to an end-to-end solution. More importantly it hampers integration of clustering as a module in a larger end-to-end deep learning system. In contrast to these methods, we show how one can jointly train a deep representation and cluster centroids with the standard $k$-means objective using backpropagation and conventional deep learning optimisers.
\section{Concrete $k$-Means \label{methodology}}
We first introduce the conventional $k$-means model. Following this, we show how to adapt the $k$-means objective to train cluster centroids and a deep neural network representation simultaneously. We refer to this novel generalisation as Concrete $k$-Means (CKM), due to the use of the Concrete distribution~\cite{maddison2016concrete}.

\subsection{Conventional $k$-Means}
The $k$-means algorithm groups data points, $\{\vec x_i\}_{i=1}^N$ from some space, $\mathcal{X} \subset \mathbb{R}^d$, into $k$ different clusters parameterised by centroids, $\{\vec \mu_i\}_{i=1}^k$, also from $\mathbb{R}^d$. By stacking each $\vec \mu_i$, the centroids can be collectively represented as a matrix, $M \in \mathbb{R}^{k\times d}$, where each row corresponds to a cluster centre. The $k$-means objective is to find the assignment and set of centroids that minimise the distance between each point and its associated centroid. 

\begin{align}
    \label{eq:k_means_obj}
    &\min_{H,M} \|X - HM\|^2_{F} \\
    \text{s.t.} & \|\vec h_j\|_1 = 1, H \in \{0, 1\}^{N\times k} \nonumber
\end{align}

\noindent where $H \in \{0, 1\}^{N\times k}$ is a binary matrix that represents the cluster assignments of each point, $\vec h_j$ is the $j$th row of $H$, and $N$ is number of data points.

The most common method for learning $k$-means clusters is Lloyd's algorithm~\cite{lloyd1982least}, which can be formalised as an alternating optimisation problem. The first step is to find the optimal cluster assignments given the current cluster centres,
\begin{align}
    \label{eq:k_means_step_1}
    &\min_{H} \|X - HM\|^2_{F} \\
    \text{s.t.} & \|\vec h_j\|_1 = 1, H \in \{0, 1\}^{N\times k} \nonumber
\end{align}
 The second step is to optimise with respect to the cluster centres keeping assignments fixed,
\begin{equation}
    \label{eq:k_means_step_2}
    \min_{M} \|X - HM\|^2_{F}.
\end{equation}

Both of these optimisation problems permit closed form solutions: Equation~\ref{eq:k_means_step_1} can be solved by finding the cluster centre closest to each data point, and Equation~\ref{eq:k_means_step_2} is minimised when each cluster centre is set to the mean of its assigned data points. Lloyd's algorithm alternates between finding locally optimal solutions to these two problems until the cluster assignments become stable.

\subsection{Deep $k$-Means with Concrete Gradients\label{v_dk_means}}
\subsubsection{A Regularised Deep Embedding Space}
Deep $k$-means strategies aim to cluster the data in a learned embedding space $\mathcal{Z}$ rather than the raw input space $\mathcal{X}$. The embedding is defined via a learned neural network $\mathbf{z}=f_{\vec \phi}(\mathbf{x})$, and we will train it to  support $k$-means clustering better than the original space. Following previous work \cite{xie2016unsupervised,yang2017towards}, we avoid degenerate solutions by defining an autoencoder that regularizes the latent space by reconstructing the original input. Specifically, we define an encoder, $f_{\vec \phi}: \mathcal{X} \rightarrow \mathcal{Z}$, which maps from the input space to the latent space, and decoder, $g_{\vec \varphi}: \mathcal{Z} \rightarrow \mathcal{X}$, which maps from the latent space back to the input space. These networks are then composed and their parameters, $\vec \phi$ and $\vec \varphi$, are trained to minimise the reconstruction error,
\begin{equation}
    \mathcal{L}^{AE}(X, \vec \phi, \vec \varphi) = \sum_{i=1}^N \|\vec x_i - g_{\vec \varphi}(f_{\vec \phi}(\vec x_i))\|_2^2.
\end{equation}

\begin{figure}[t]
\centering
\resizebox{0.5\columnwidth}{!}{
\input{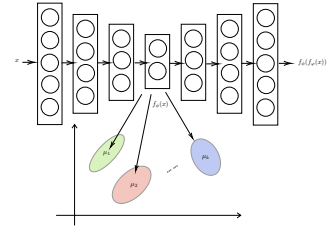}
}
\caption{Illustration of the of Concrete $k$-means architecture. An input item $\vec x$ is embedded by $f_{\vec \phi}$, and clusters $\vec \mu$ are learned in this low dimensional latent space. Decoder $g_{\vec \varphi}$ regularizes the latent space.}
\label{diagram_concrete_kmeans}
\end{figure}

\subsubsection{Differentiable Clustering in the Latent Space}

The proposed algorithm performs clustering in the latent space $\mathcal{Z}$ rather than the input space $\mathcal{X}$. In the conventional $k$-means algorithm, a data point is assigned to the cluster with the nearest centroid, as measured by Euclidean distance. The hard assignment operation is not differentiable, thus precluding the direct use of standard gradient-based optimisation techniques for training neural networks. In order to perform both the hard assignment operation and obtain gradient estimates for training he $k$-means objective, we borrow the idea of Straight-Through Gumbel-Softmax~\cite{jang2016categorical}. This reparameterisation trick enables the use of a probabilistic hard assignment during the forward propagation, while also allowing gradients to be backpropagated through a soft assignment in order to train the network. We keep the Euclidean distance of the traditional $k$-means algorithm, and model cluster assignment probabilities using normalised radial basis functions (RBFs),
\begin{equation}
\label{rbf}
p(C_{i,j}|\vec x_i) = \frac{\text{exp}\{-\sigma^{-2}\|\vec z_i - \vec \mu_j\|_2^2\}}{\sum_{c=1}^{k}\text{exp}\{-\sigma^{-2}\|\vec z_i - \vec \mu_c\|_2^2\}},
\end{equation}
\noindent where $C_{i,j}$ is the event that instance $i$ is assigned to cluster $j$, and $\vec z_i = f_{\vec \phi}(\vec x_i)$, and we have omitted the dependence of $p$ on the cluster centres, $M$, and network parameters, $\vec \phi$, to keep notation compact. We would like to draw samples represented as one-hot vectors from $p(C_i | \vec x_i)$, while simultaneously being able to backpropagate through the sampling process. This can be accomplished by instead sampling from a Gumbel-Softmax distribution---a continuous relaxation of the distribution of one-hot encoded samples from $p(C_i | \vec x_i)$. By introducing Gumbel distributed random variables, $G$, one can make use of a reparameterisation trick to sample from the Gumbel-Softmax distribution,
\begin{equation}
\label{reparameterization}
h_{i,j} = \frac{\text{exp}\{\tau^{-1}(\text{log}(p(C_{i,j}|\vec x_i)) + G_j)\}}{\sum^{k}_{c=1}\text{exp}\{\tau^{-1}(\text{log}(p(C_{i,c}|\vec x_i)) + G_c)\}},
\end{equation}
where $h_{i,j}$ is the $j$th component of the vector, $\vec h_i$ corresponding to instance $\vec x_i$, and  $\tau \in (0,\infty)$ is a temperature hyperparameter used for controlling the entropy of the continuous relaxation. As $\tau$ goes to zero, $\vec h_i$ converges towards true one-hot samples from $p(C_i | \vec x_i)$. In contrast, as $\tau$ goes to infinity, the $\vec h_i$ converge towards a uniform distribution. In practice, we start training with a high temperature and gradually anneal it towards zero as training progresses.

During test time, the $\argmax$ of $p(C_i | \vec x_i)$ is taken, rather than sampling via $\vec h_i$. The $\vec h_i$ vectors can be discretised by rounding the largest component to one, and all others to zero, giving a truly discrete sample distributed according to $p(C_i | \vec x_i)$. We denote the discretization of $\vec h_i$ by $\Tilde{\vec h}_i$. With this notation, we define the concrete $k$-means loss as
\begin{equation}
\mathcal{L}^{CKM}(X, M, \vec \phi) = \sum_{i=1}^N ||f_{\vec \phi}(\vec x_i)- \Tilde{\vec h}_i M||^2_2,\label{eq:ckmmain}
\end{equation}
noting that $\Tilde{\vec h}_i$ is a row vector. During the forward propagation,  $\Tilde{\vec h}_i$ is used for evaluating the $k$-means loss. During the backward pass, the gradient is estimated by back-propagating though the same loss, but parameterised by $\vec h_i$ instead of $\Tilde{\vec h}_i$. This method of computing gradients for one-hot encoded categorical variables is known as the straight through Gumbel-softmax estimator~\cite{jang2016categorical}, or the concrete estimator~\cite{maddison2016concrete}.  

\subsubsection{Summary}
\begin{figure}[t]
\centering
\resizebox{0.9\columnwidth}{!}{
\tikzset{every picture/.style={line width=0.75pt}} 

\begin{tikzpicture}[x=0.75pt,y=0.75pt,yscale=-1,xscale=1]

\draw   (70.5,383.92) -- (111.5,383.92) -- (111.5,421.92) -- (70.5,421.92) -- cycle ;
\draw    (27.5,399.92) -- (61,399.92) ;
\draw [shift={(63,399.92)}, rotate = 180] [color={rgb, 255:red, 0; green, 0; blue, 0 }  ][line width=0.75]    (10.93,-3.29) .. controls (6.95,-1.4) and (3.31,-0.3) .. (0,0) .. controls (3.31,0.3) and (6.95,1.4) .. (10.93,3.29)   ;

\draw   (170,380.92) -- (237.5,380.92) -- (237.5,423.92) -- (170,423.92) -- cycle ;
\draw    (122.5,399.92) -- (162,399.92) ;
\draw [shift={(164,399.92)}, rotate = 180] [color={rgb, 255:red, 0; green, 0; blue, 0 }  ][line width=0.75]    (10.93,-3.29) .. controls (6.95,-1.4) and (3.31,-0.3) .. (0,0) .. controls (3.31,0.3) and (6.95,1.4) .. (10.93,3.29)   ;

\draw    (270.5,312.92) -- (208.91,374.51) ;
\draw [shift={(207.5,375.92)}, rotate = 315] [color={rgb, 255:red, 0; green, 0; blue, 0 }  ][line width=0.75]    (10.93,-3.29) .. controls (6.95,-1.4) and (3.31,-0.3) .. (0,0) .. controls (3.31,0.3) and (6.95,1.4) .. (10.93,3.29)   ;

\draw   (282,382.92) -- (426.5,382.92) -- (426.5,422.92) -- (282,422.92) -- cycle ;
\draw    (242.5,400.92) -- (276,400.92) ;
\draw [shift={(278,400.92)}, rotate = 180] [color={rgb, 255:red, 0; green, 0; blue, 0 }  ][line width=0.75]    (10.93,-3.29) .. controls (6.95,-1.4) and (3.31,-0.3) .. (0,0) .. controls (3.31,0.3) and (6.95,1.4) .. (10.93,3.29)   ;

\draw   (456,382.92) -- (521.5,382.92) -- (521.5,423.92) -- (456,423.92) -- cycle ;
\draw    (427.5,403.92) -- (451,403.92) ;
\draw [shift={(453,403.92)}, rotate = 180] [color={rgb, 255:red, 0; green, 0; blue, 0 }  ][line width=0.75]    (10.93,-3.29) .. controls (6.95,-1.4) and (3.31,-0.3) .. (0,0) .. controls (3.31,0.3) and (6.95,1.4) .. (10.93,3.29)   ;

\draw    (295.5,311.92) -- (478.61,375.27) ;
\draw [shift={(480.5,375.92)}, rotate = 199.07999999999998] [color={rgb, 255:red, 0; green, 0; blue, 0 }  ][line width=0.75]    (10.93,-3.29) .. controls (6.95,-1.4) and (3.31,-0.3) .. (0,0) .. controls (3.31,0.3) and (6.95,1.4) .. (10.93,3.29)   ;

\draw    (299.5,463.92) -- (315.65,429.73) ;
\draw [shift={(316.5,427.92)}, rotate = 475.28] [color={rgb, 255:red, 0; green, 0; blue, 0 }  ][line width=0.75]    (10.93,-3.29) .. controls (6.95,-1.4) and (3.31,-0.3) .. (0,0) .. controls (3.31,0.3) and (6.95,1.4) .. (10.93,3.29)   ;

\draw    (534.5,403.92) -- (565,403.92) ;
\draw [shift={(567,403.92)}, rotate = 180] [color={rgb, 255:red, 0; green, 0; blue, 0 }  ][line width=0.75]    (10.93,-3.29) .. controls (6.95,-1.4) and (3.31,-0.3) .. (0,0) .. controls (3.31,0.3) and (6.95,1.4) .. (10.93,3.29)   ;

\draw   (571,382.92) -- (643.5,382.92) -- (643.5,423.92) -- (571,423.92) -- cycle ;
\draw  [dash pattern={on 4.5pt off 4.5pt}]  (165.5,391.92) -- (119,391.92) ;
\draw [shift={(117,391.92)}, rotate = 360] [color={rgb, 255:red, 0; green, 0; blue, 0 }  ][line width=0.75]    (10.93,-3.29) .. controls (6.95,-1.4) and (3.31,-0.3) .. (0,0) .. controls (3.31,0.3) and (6.95,1.4) .. (10.93,3.29)   ;

\draw [color={rgb, 255:red, 223; green, 27; blue, 27 }][dash pattern={on 4.5pt off 4.5pt}]  (275.5,392.92) -- (240,392.92) ;
\draw [shift={(238,392.92)}, rotate = 360] [color={rgb, 255:red, 223; green, 27; blue, 27 }  ][line width=0.75]    (10.93,-3.29) .. controls (6.95,-1.4) and (3.31,-0.3) .. (0,0) .. controls (3.31,0.3) and (6.95,1.4) .. (10.93,3.29)   ;

\draw  [dash pattern={on 4.5pt off 4.5pt}]  (566.5,393.92) -- (531,393.92) ;
\draw [shift={(529,393.92)}, rotate = 360] [color={rgb, 255:red, 0; green, 0; blue, 0 }  ][line width=0.75]    (10.93,-3.29) .. controls (6.95,-1.4) and (3.31,-0.3) .. (0,0) .. controls (3.31,0.3) and (6.95,1.4) .. (10.93,3.29)   ;

\draw  [dash pattern={on 4.5pt off 4.5pt}]  (454.5,394.92) -- (427,394.92) ;
\draw [shift={(425,394.92)}, rotate = 360] [color={rgb, 255:red, 0; green, 0; blue, 0 }  ][line width=0.75]    (10.93,-3.29) .. controls (6.95,-1.4) and (3.31,-0.3) .. (0,0) .. controls (3.31,0.3) and (6.95,1.4) .. (10.93,3.29)   ;

\draw  [dash pattern={on 4.5pt off 4.5pt}]  (477.5,365.92) -- (301.39,305.57) ;
\draw [shift={(299.5,304.92)}, rotate = 378.91999999999996] [color={rgb, 255:red, 0; green, 0; blue, 0}  ][line width=0.75]    (10.93,-3.29) .. controls (6.95,-1.4) and (3.31,-0.3) .. (0,0) .. controls (3.31,0.3) and (6.95,1.4) .. (10.93,3.29)   ;

\draw  [color={rgb, 255:red, 223; green, 27; blue, 27 }][dash pattern={on 4.5pt off 4.5pt}]  (204.5,366.92) -- (268.09,303.34) ;
\draw [shift={(269.5,301.92)}, rotate = 495] [color={rgb, 255:red, 223; green, 27; blue, 27 }  ][line width=0.75]    (10.93,-3.29) .. controls (6.95,-1.4) and (3.31,-0.3) .. (0,0) .. controls (3.31,0.3) and (6.95,1.4) .. (10.93,3.29)   ;

\draw (18,398.92) node   {$\vec x$};
\draw (91,402.92) node   {$f_{\vec \phi}(\vec x)$};
\draw (203.75,402.42) node   {$p(C|\vec x)$};
\draw (284,298.92) node   {$M$};
\draw (354.25,402.92) node   {$\text{ST-GumbelSoftmax}(\cdot)$};
\draw (488.75,403.42) node   {$\Tilde{\vec h}\cdot M$};
\draw (338,477) node   {$G\ \sim \text{Gumbel}$};
\draw (607.25,403.42) node   {$\mathcal{L}^{CKM}$};

\end{tikzpicture}
}
\caption{A computational graph view of the information flow for the concrete $k$-means algorithm. Solid arrows indicate computation during forward propagation, and dashed arrows indicate gradient flow during backpropagation. The red dashed arrows show which gradients are computed by the concrete gradient estimator.}
\label{deep_k-meanscg}
\end{figure}
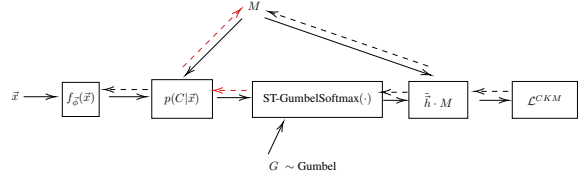

To train our Concrete $k$-means, we optimize the main CKM objective in Eq.~\ref{eq:ckmmain} along with the autoencoder, with respect to encoder and decoder parameters as well as cluster centres. The full objective is:
\begin{equation}
   \min_{M, \vec \phi, \vec \varphi} \mathcal{L}^{AE}(X, \vec \phi, \vec \varphi) + \lambda_{1} \mathcal{L}^{CKM}(X, M, \vec \phi),
\label{overall}
\end{equation}
where $\lambda_1$ is a regularisation strength hyperparameter. The stochastic computational graph \cite{schulman2015gradient} in Figure~\ref{deep_k-meanscg} illustrates the flow of information during training for both the forward and backward passes. Dashed arrows indicate the flow of gradients, and solid arrows are activations computed during the forward propagation. The red dashed arrows represent the gradients estimated by our method that would typically be blocked by hard assignment or generated by soft assignment in other methods.

In practice, pretraining the feature extractor using the autoencoder reconstruction before jointly training the full objective improves the final clustering solution. The algorithm and architecture for training deep CKM are outlined in Algorithm~\ref{ck_means} and Figure~\ref{diagram_concrete_kmeans} respectively. 

\begin{algorithm2e}[t]
\label{ck_means}
\SetKwProg{input}{Input:}{}{}
\SetKwProg{output}{Onput:}{}{}
\SetKwProg{init}{Init:}{}{}
\SetAlgoLined
\input{$X, \alpha, \eta ,\lambda$}{}
\output{$f_{\vec \phi}, g_{\vec \varphi}, M$}{}
\Begin{
\init{$\vec \phi, \vec \varphi$}{}
\While{not converge}{
 $\vec \phi \leftarrow \vec \phi - \alpha\nabla_{\vec \phi} \mathcal{L}^{AE}(X) $\;
 $\vec \varphi \leftarrow \vec \varphi - \alpha\nabla_{\vec \varphi} \mathcal{L}^{AE}(X) $\;
 } 
\init{$M \text{ with }  k\text{-means}^{++}$}{}
\While{not converge}{
 $\vec \phi \leftarrow \vec \phi - \eta\nabla_{\vec \phi} (\mathcal{L}^{AE}(X)+ \lambda \mathcal{L}^{CKM}(X))$\;
 $\vec \varphi \leftarrow \vec \varphi - \eta\nabla_{\vec \varphi}\mathcal{L}^{AE}(X)$\;
 $M \leftarrow M - \eta\nabla_{M} \mathcal{L}^{CKM} (X)$\;
}
}
\caption{Concrete $k$-means clustering}
\end{algorithm2e}

\subsubsection{Shallow Concrete $k$-means}
Our algorithm is motivated by the vision of joint clustering and representation learning. Nevertheless, it is worth noting that as a byproduct it provides a novel optimisation strategy for the conventional $k$-means objective in  Equation~\ref{eq:k_means_obj}. We simply run CKM on raw features, which can be interpreted as fixing the encoder and decoder to the identity function, and solve Equation~\ref{overall} for centroids $M$ alone. Thus we use stochastically estimated gradients to solve conventional $k$-means by gradient descent rather than alternating minimisation \cite{lloyd1982least}.
\section{Experiments}
In this section, we evaluate CKM in conventional shallow and deep clustering.

\subsection{Shallow Clustering}
The concrete $k$-means method presented in Section~\ref{v_dk_means} does not require the presence of a feature extraction network, and can thus be used to optimise the $k$-means objective in the `shallow' setting where Lloyd's \cite{lloyd1982least} and $k$-means++ \cite{arthur2017kmeansplus} are typically applied. Our first experiment aims to confirm if the CKM gradient-based stochastic optimisation matches the performance of the standard $k$-means solvers. Table~\ref{shallow_ckm} reports the clustering results of our shallow CKM and sklearn's $k$-means++ implementation on ten UCI datasets. The evaluation metrics used for these experiments are normalized mutual information (NMI) \cite{cai2010locally}, adjusted rand index (ARI) \cite{yeung2001details}, and cluster purity (ACC). The values of ACC and NMI are rescaled to lie between zero and one, with higher values indicating better performance. The range of the ARI is negative one to one. We can see that CKM performs comparably to the standard $k$-means optimizer.

\begin{table*}[t]
\small
\begin{center}
\begin{tabular}{l@{\hskip 0.8cm} c c c@{\hskip 0.8cm} c c c }
\toprule
 & \multicolumn{3}{c}{Shallow CKM} & \multicolumn{3}{c}{$k$-means++ 
 }\\
 & NMI & ARI & ACC & NMI & ARI & ACC \\
 \midrule
pendigits&0.50$\pm$0.04&0.33$\pm$0.05&0.49$\pm$0.05&0.51$\pm$0.04&0.34$\pm$0.05&0.49$\pm$0.05 \\
dig44&0.33$\pm$0.04&0.20$\pm$0.05&0.40$\pm$0.05&0.33$\pm$0.04&0.20$\pm$0.05&0.40$\pm$0.05 \\
vehicle&0.15$\pm$0.03&0.09$\pm$0.03&0.40$\pm$0.03&0.15$\pm$0.03&0.09$\pm$0.03&0.40$\pm$0.03 \\
letter&0.35$\pm$0.01&0.13$\pm$0.01&0.26$\pm$0.01&0.35$\pm$0.01&0.13$\pm$0.01&0.25$\pm$0.01 \\
segment&0.41$\pm$0.05&0.27$\pm$0.05&0.46$\pm$0.04&0.41$\pm$0.05&0.27$\pm$0.05&0.46$\pm$0.04 \\
waveform&0.35$\pm$0.04&0.27$\pm$0.04&0.57$\pm$0.05&0.36$\pm$0.01&0.25$\pm$0.01&0.52$\pm$0.02 \\
vowel&0.41$\pm$0.01&0.21$\pm$0.01&0.36$\pm$0.02&0.42$\pm$0.01&0.21$\pm$0.01&0.36$\pm$0.02 \\
spambase&0.10$\pm$0.03&0.09$\pm$0.05&0.66$\pm$0.04&0.10$\pm$0.03&0.09$\pm$0.05&0.66$\pm$0.04 \\
twonorm&0.84$\pm$0.00&0.91$\pm$0.00&0.98$\pm$0.00&0.84$\pm$0.01&0.91$\pm$0.01&0.98$\pm$0.00 \\
sat&0.58$\pm$0.05&0.48$\pm$0.08&0.64$\pm$0.07&0.58$\pm$0.05&0.48$\pm$0.08&0.64$\pm$0.07 \\
\bottomrule
\end{tabular}
\end{center}
\caption{Shallow CKM uses gradient estimation to solve the standard fixed-feature $k$-means problem equally well to the conventional alternating minimisation based $k$-means++ \cite{arthur2017kmeansplus} implemented in scikit-learn.}
\label{shallow_ckm}
\end{table*}

\subsection{Deep Clustering \label{pure_clustering}}
\subsubsection{Datasets and Settings}
\textbf{Datasets} We conduct deep clustering experiments are using the following datasets from the image and natural language domains: \textbf{MNIST}\quad \cite{lecun1998gradient} consists of 70,000 greyscale images of handwritten digits. There are 10 classes and each image is $28 \times 28$ pixels, with the digits appearing inside the central $20 \times 20$ pixel area. \textbf{USPS}\quad is a dataset of $16 \times 16$ pixel handwritten digit images. The first 7,291 images are designated as the training fold, and the remaining 2,007 are used for evaluating the final performance of the models. \textbf{20Newsgroups}\quad was generated by collecting a total of 18,846 posts over 20 different newsgroups. We use the same preprocessing as~\cite{yang2017towards}, where the tf-idf representation of the 2,000 most frequently occurring words are used as features.

\textbf{Architecture} Like most deep clustering methods (e.g., \cite{yang2017towards} and \cite{xie2016unsupervised}), our approach involves pretraining an autoencoder before optimising the clustering objective. The encoder architecture used for the clustering experiments on MNIST and USPS contains four fully connected layers with 500, 500, 2000, and 10 units, respectively. For the 20Newsgroup experiments, the smaller encoder with 250, 100, and 20 units described by~\cite{yang2017towards} is used. The decoder that maps the hidden representation back to the input space is the mirror version of the encoder.

\textbf{Competitors} Comparisons are made with \textbf{DEC}~\cite{xie2016unsupervised} and \textbf{DCN}~\cite{yang2017towards}, as well as some simple baselines. \textbf{KM} applies classic shallow K-means  to raw input features from $\mathcal{X}$. \textbf{AE+KM} performs two step dimensionality reduction and clustering by training an autoencoder with the same architecture as CKM to embed instances into the latent space $\mathcal{Z}$, and then fixes this space before applying classic $k$-means clustering.

\subsubsection{Results}
\small
\begin{table*}[t]
\begin{adjustbox}{max width=\textwidth}
\centering
\begin{tabular}{l@{\hskip 0.8cm} c c c@{\hskip 0.8cm} c c c@{\hskip 0.8cm} c c c }
\toprule
\multirow{2}{4em}{ Method } & \multicolumn{3}{c}{MNIST} & \multicolumn{3}{c}{USPS} & \multicolumn{3}{c}{20NEWSGROUP}\\
 & NMI & ARI & ACC & NMI & ARI & ACC & NMI & ARI & ACC\\\midrule
 
KM$^h$            & 51.8$\pm$0.4 & 36.5$\pm$0.4 & 53.3$\pm$0.5 
              & 60$\pm$ 0.7  & 44 $\pm$0.9  & 58 $\pm$0.9 
              & 22.7$\pm$1.7 & 8.0$\pm$1.4  & 22.6$\pm$2.1\\
AE+KM$^h$         & 74.3$\pm$0.9 & 66.9$\pm$0.8 & 80.6$\pm$1.2
              & 68.1$\pm$0.3 & 59.4$\pm$0.4 & 68.4$\pm$0.7
              & 42.0$\pm$1.7 & 28.3$\pm$1.2 & 44.3$\pm$2.3\\
DEC$^e$           & 80.4$\pm$1.3 & 76.3$\pm$1.8 & 84.2$\pm$1.7
              & \textbf{72.6$\pm$1.1} & \textbf{63.8$\pm$0.9} & 71.1$\pm$2.5
              & \textbf{48.6$\pm$1.2} & \textbf{35.4$\pm$1.4} & \textbf{49.1$\pm$2.5}\\
DCN$^h$         
              & \textbf{81.7$\pm$1.1} & 75.2$\pm$1.2 & 83.1$\pm$1.9
              & 71.9$\pm$1.2 & 61.9$\pm$1.4 & \textbf{73.9$\pm$0.8}
              & 44.7$\pm$1.5 & 34.4$\pm$1.3 & 46.3$\pm$2.9\\
\midrule
CKM$^{h,e}$           & {81.4$\pm$1.8} &\textbf{77.7$\pm$1.1}& \textbf{85.4$\pm$2.1} 
              & 70.7$\pm$0.2       & 61.3$\pm$0.2  &  72.1$\pm$0.4
              & 46.5$\pm$1.4       & 34.1$\pm$1.6   & 47.3$\pm$2.3\\
\bottomrule
\end{tabular}
\end{adjustbox}
\caption{Deep clustering results on MNIST, USPS and 20 Newsgroups. $^h$ indicates methods with interpretable hard assignments, and $^e$ indicates methods with end-to-end learning by backpropagation. Only our Concrete $k$-means combines hard assignment and end-to-end learning.}
\label{over_all}
\end{table*}

In Table~\ref{over_all} we illustrate the results of our CKM and the comparison with other models. For all experiments in this section, $k$ is set to the number of classes present in the dataset. We run each method 15 times with different initial random seeds and report the mean and standard deviation of each result. 

From the results, we can see that all the deep methods outperform shallow $k$-means on raw-features (KM), and furthermore all the jointly trained methods outperform the two-step baseline (AE+KM). Compared to the published state of the art methods, our CKM approach generally performs best on MNIST, and comparably to DEC and DCN on USPS and 20Newsgroup. Importantly, our CKM is the only high-performing method to combine the favorable properties of hard-assignment, which is important for interpretability in many applications \cite{kearns1997information}; and end-to-end deep learning, which is important to be able to integrate clustering functionality as a module into a larger backpropagation-driven system.

\textbf{Runtime Efficiency:} Comparing the three deep clustering methods, DCN's alternating optimisation is slower than the end-to-end DEC and CKM. For MNIST, the clock time per epoch is 11$s$, 10$s$, and 36$s$ for CKM, DEC and DCN respectively.
\section{Conclusion}
This paper proposes the concrete $k$-means deep clustering framework. Our stochastic hard assignment method is able to estimate gradients of the nondifferentiable $k$-means loss function with respect to cluster centres. This, in turn, enables end-to-end training of a neural network feature extractor and a set of cluster centroids in this latent space. Our experimental results show that the proposed method is competitive with state-of-the-art approaches for solving deep clustering problems.
\bibliographystyle{IEEEbib}
\bibliography{strings}

\end{document}